\documentclass{article}

\usepackage{url}
\usepackage{algorithm}
\usepackage{algorithmic}
\usepackage{subfigure}
\usepackage[T1]{fontenc}
\usepackage{amsfonts}
\usepackage{graphicx}
\usepackage{caption}

\begin{document}

\title{Improving Genetic Algorithms Performance via Deterministic Population Shrinkage} 
\author{Juan Luis Jim\'enez Laredo\thanks{\texttt{Corresponding author: juanlu@ugr.es} Department of Architecture and Computer Technology, University of Granada, Spain} \and 
Carlos Fernandes\thanks{LASEEB-ISR/IST. University of Lisbon, Portugal} \and 
Juan Juli\'an Merelo\thanks{Department of Architecture and Computer Technology, University of Granada, Spain} \and
Christian Gagn\'e\thanks{Laboratoire de vision et syst\`emes num\'eriques, Laval University, Canada}}
\date{}

\maketitle

\begin{abstract}

Despite the intuition that the same population size is not needed throughout the run of an Evolutionary Algorithm (EA), most EAs use a fixed population size.
This paper presents an empirical study on the possible benefits of a Simple Variable Population Sizing (SVPS) scheme on the performance of Genetic Algorithms (GAs).
It consists in decreasing the population for a GA run following a predetermined schedule, configured by a speed and a severity parameter.
The method uses as initial population size an estimation of the minimum size needed to supply enough building blocks, using a fixed-size selectorecombinative GA converging within some confidence interval toward good solutions for a particular problem.
Following this methodology, a scalability analysis is conducted on deceptive, quasi-deceptive, and non-deceptive trap functions in order to assess whether SVPS-GA improves performances compared to a fixed-size GA under different problem instances and difficulty levels.
Results show several combinations of speed-severity where SVPS-GA preserves the solution quality while improving performances, by reducing the number of evaluations needed for success.
\end{abstract}

\section*{Note:}
This paper is a pre-print version of the following paper that you can cite as:\\
\textbf{Juan Luis J. Laredo, Carlos Fernandes, Juan Julián Merelo, and Christian Gagné. 2009. Improving genetic algorithms performance via deterministic population shrinkage. In Proceedings of the 11th Annual conference on Genetic and evolutionary computation (GECCO '09). Association for Computing Machinery, New York, NY, USA, 819–826. \url{https://doi.org/10.1145/1569901.1570014}}

\section{Introduction}

Setting an adequate population size is a key to obtain good performances
in a Genetic Algorithm (GA), that is, to preserve a good quality in
the solutions without spending extra computational efforts. That way, a
small problem instance will require a smaller population size than a
larger instance of a more difficult problem
\cite{Goldberg92geneticalgorithms,Harik97thegambler}. Additionally,
improving the GA performance is also possible by varying the
population size during the GA run, adapting it as the algorithm is converging
toward some solutions \cite{plagues, laredo08:churn}. 

Population sizing theory, based on Goldberg's facetwise
decomposition for designing competent GAs \cite{goldberg:competent},
focuses on determining population sizing according to problem difficulty.
Despite the issue that has been pointed out in Lobo and Lima's review of
adaptive population sizing schemes \cite{lobo07:review}, such a theory
has not received much attention with respect to variable population sizing
schemes. There are just a few studies that have taken it into account, with as
prime example population sizing through estimation of schema variances by
Smith and Smuda \cite{smith:resizing}, with selection errors probability and 
adjustment made according to an expected selection loss provided by the user.

Therefore, the aim of this paper is to provide a general framework for
evaluating performance of variable population sizing schemes from the
population sizing theory perspective. For that purpose, a method based on
bisection \cite{sastry:bisection} is used to estimate the minimum
size of the initial population $P_{init}$ required to supply enough building blocks
(BBs) so that a fixed-size selectorecombinative GA\footnote{The assumption of a
selectorecombinative GA (without mutation) is made as the only
source of diversity is then the initial supply of BBs.} will converge toward
optimal solutions. That population size is then used for bootstrapping another selectorecombinative
GA, which is using for the tested variable population sizing scheme. This allows assessment of the solutions quality and
computational effort required. Given that both fixed and variable
sizing schemes are provided with the same initial supply of BBs, some conclusions
can be drawn from the difference of performances. Additionally, a
scalability analysis is performed on different instances and
complexities of decomposable trap functions \cite{ackley:trap}. Hence,
it is possible not only to analyze a given performance, but also, to analyze how
it is affected by problem size and complexity. 

We propose to test with a simple variable population sizing scheme (SVPS). The SVPS is a deterministic function that monotonically reduces the population size during the GA run. The function slope can be controlled by setting two parameters, one for controlling the shrinking speed and the other for controlling the severity of population resizing. The intuition justifying the approach is that, at an early stage of a GA run, larger population sizes produce a wider exploration of the search landscape. Meanwhile, smaller population sizes increase the exploitation of promising regions as the GA converges  \cite{lobo07:review}. 

Nevertheless, our intent in the current work is to address the following points:
\begin{enumerate}
\item Provide a general framework to test different variable population sizing schemes from the population sizing theory perspective. Following the same steps, SVPS could be replaced by new or suitable schemes in the literature.
\item Check whether variable population sizing schemes in GAs are able to outperform an equivalent parameterized GA using a fixed size scheme. If the SVPS-GA improves the fixed-size GA performance, that would mean that there are better schemes for population sizing than the standard fixed size scheme.
\item Gain some insight into the dynamics of the SVPS scheme. The idea of SVPS is to maintain a good quality in the solutions using less evaluations. Nevertheless, the seamless run of a GA makes it difficult to estimate the desired population size decreasing. Additionally, a given strategy could be more or less adequate to different problem instances or problem complexities.
\end{enumerate}

The rest of the paper is organized as follows: Section \ref{sec:related} presents the state of the art in variable population sizing schemes. The proposed methodology is described in  Section \ref{sec:methodology} and the experimental setup in Section \ref{sec:setup}. Section \ref{sec:results} explores the dynamics of the SVPS, showing that a variable population sizing scheme can yield a better performance than a fixed population sizing scheme. Finally, results are discussed in Section \ref{sec:conclusions}.

\section{Related Works}
\label{sec:related}

This section provides a brief description of some other varying population methods previously proposed in the Evolutionary Computation research field. Following the classification of parameter control mechanisms given by Eiben \emph{et al.} \cite{Hinterding99parametercontrol}, we may say that some of the techniques described below fall into the adaptive methods categories (GAVaPS, for instance), while others, like RVPS \cite{costa99variation} and PRoFIGA \cite{Eiben04evolutionaryalgorithms} are deterministic methods. Our proposal may be classified as deterministic, because the population size in each generation is defined in the beginning of the search by two parameters; the size is always forced to decrease, with more or less speed, and ends with more or less individuals, depending on those two parameters. Other methods follow a different policy and adapt the size during the run according to the state of the search. One of those algorithms, proposed in 1994 by Arabas \emph{et al.} \cite{arabas94gavaps}, is the Genetic Algorithm with Varying Population Size (GAVaPS).     

GAVaPS does not hold an explicit selection mechanism. As in natural systems, population size is defined by the birth and death of individuals occurring at each iteration. A parameter called lifetime is introduced. It defines the number of generations in which each individual is allowed to remain alive. After its creation, the chromosome is assigned to a specific lifetime, according to its fitness. Three lifetime calculation methods are proposed. The algorithm proceeds in a generational manner, at each time step increasing each individual's age. When an individual's age exceeds its lifetime, the chromosome is removed from the population. Since fittest individuals remain in the population for more generations, thus having a higher probability to be engaged in a reproduction process and generate offspring, GAVaPS' chromosomes have equal probability to be selected to reproduce, independently of their fitness value. This concept of lifetime/age provides the algorithm with the necessary selection pressure, which reduces the need for selection strategies: GAVaPS randomly pairs the chromosomes for crossover operations. The intensity of the pressure is controlled by two parameters, $\mathit{minLT}$ and $\mathit{maxLT}$, that define, respectively, the minimum and maximum lifetime allowed for each chromosome. Higher difference between the two values leads to a more selective algorithm. However, this process may have a serious drawback since increasing the $\mathit{maxLT}$ parameter will result in larger populations and, as stated above, an increasingly high population size is a characteristic of GAVaPS. The algorithm also introduces another parameter: reproduction rate ($\rho$); its value defines the number of new chromosomes created in each generation $t$, depending on the size of the current population. GAVaPS was tested for the studies presented in \cite{fernandes06srpea} and \cite{Eiben04evolutionaryalgorithms} and the results lead to the conclusion that GAVaPS is extremely sensitive to the reproduction rate, and very often the population grows exponentially or becomes extinct. 

A similar approach was attempted with The Adaptive Population size Genetic Algorithm (APGA) \cite{back00withoutparameters}. The only difference between this algorithm and GAVaPS resides in reproduction rate, which in APGA has a fixed value of two individuals. This technique follows the reproduction strategy of the Steady-State GA and prevents the population from growing out of control as it often happens with GAVaPS. On the other hand, such a low reproduction rate results in populations with few individuals unless a high value for $\mathit{maxLT}$ is used. But, even in the last case, the population size is very stable and apparently does not react to the evolution process and different search stages \cite{back00withoutparameters}. The algorithm appears to perform well on some problems and clearly outperformed GAVaPS when applied to the Spears' multi-modal problems \cite{Eiben04evolutionaryalgorithms}. However, Lobo and Lima \cite{lobo06:revisiting} questioned the results in \cite{Eiben04evolutionaryalgorithms} and proved that there is an upper bound equal to $2\mathit{maxLT}+1$ for APGA's population size, after $\mathit{maxLT}$ generations.

Eiben, Marchiori and Valk\'o proposed in \cite{Eiben04evolutionaryalgorithms}
the Population Resizing on Fitness Improvement GA (PRoFIGA). The variation process of PRoFIGA is based on the improvement of the best fitness in the population. The process intends to balance exploration and exploitation by growing the population in earlier and exploratory stages and gradually decreasing it in later stages of the search. When the population becomes trapped in local optima, the process is supposed to generate another growing phase of the population, thus increasing diversity and escaping the local optima. The authors present a heuristic for size variation during the run that increases or decreases the population size according to whether or not the best fitness of the population has been improved and, if the later case is observed, for how long it has remained unchanged. The main drawback of this algorithm are its extra six parameters, which makes it very hard to tune for a non expert user.

In the Random Variation of Population Size GA (RVPS) \cite{costa99variation} the population size is randomly changed during the run. The authors concluded that in some cases the performance of RVPS is equivalent to the standard GA. So, when there are no hints about the optimal population size for some problem, it may be appropriate to randomly set and vary the population size of the GA.

Like PRoFIGA and RVPS, the Saw-Tooth Genetic Algorithm \cite{koumousis06saw} is an example of a deterministic method used in the variation of the population size. In this algorithm the population size varies according to a predefined function with a saw-tooth shape. The authors concluded that the Saw-Tooth GA performed well on some particular test functions. However, besides a variable population size, the Saw-Tooth GA also uses a re-initialization mechanism to introduce genetic diversity in the population.

The Self-Regulated Population Size EA (SRP-EA), proposed by Fernandes and Rosa in \cite{fernandes06srpea}, follows GAVaPS guidelines but it manages to control the population size, thus avoiding the typical extinction and demographic burst observed in the dynamics of Arabas' algorithm. SRP-EA self-controls the population size (indirectly) via genetic diversity. There is a threshold value that adapts during the run and that defines the Hamming distance value above which two chromosomes are allowed to crossover and generate offspring and like GAVaPS the individuals are provided with a lifetime that defines the of generations that they are allowed to remain in the population. The algorithm was compared with APGA and CHC \cite{chc91} and outperformed both in the proposed test set.

Finally, there could be other reasons to use variable population sizing schemes. Laredo \emph{et al.} expose in \cite{laredo08:p2pruntime} the case of a fully distributed EA  in which the individuals have to decide on their own state of reproduction without any central control, using instead estimations about the global
population state for decision making. The population size varies at run-time as a consequence of such a
decentralized reproduction and a self-adjusting mechanism based on autonomous selection \cite{eiben:autonomous} tries to keep it stable.

Our strategy, described in the next section, does not aim at adapting the population size or varying it deterministically according to the state of the search. Alternatively, and although it may be classified as a deterministic scheme, together with PROFIGA and the Saw-Tooth GA, SVPS tries to explore the premise that states that larger populations are needed in the beginning of the search, while towards the end the GA can manage to converge with a smaller population \cite{lobo07:review}.

\section{Methodology}
\label{sec:methodology}

The methodology followed takes into account Goldberg's facetwise decomposition for designing competent GAs \cite{goldberg:competent} in order to answer whether a GA using varying population size improves a fixed-size GA.

The proposed method consists in the following three steps:
\begin{enumerate}
\item Bisection method, to estimate size $n'$ of population $P_{init}'$;
\item Refinement of size of the population size, to obtain the necessary supply of BBs;
\item Simple variation of the population size using a predetermined schedule.
\end{enumerate}
The two first steps are used to estimate the minimum initial population size ($P_{init}=\{ind_{1},ind_{2},\ldots,ind_{n}\}$) required to supply enough BBs for a reliable convergence to the problem optimum in a fixed-size population GA. 

In the third step, $P_{init}$ is used as the initial population of the SVPS-GA in which the population decreases according to a parameterized speed ($\tau$) and severity ($\rho$). Any combination $\tau$-$\rho$ has to preserve the quality of solutions while improving the number of evaluations.

Steps 1, 2, and 3 are exposed in sections \ref{sec:bisection}, \ref{sec:refinement} and \ref{sec:varying}, respectively. Note that the SVPS-GA (step 3) could be changed for another variable, adaptive or self-adaptive population sizing scheme. Therefore, this methodology provides a framework to test different population resizing schemes. 

\subsection{Bisection Method}
\label{sec:bisection}

The bisection method \cite{sastry:bisection} estimates
the optimal population size $n'$ to solve a problem instance, that is, the lowest $n'$ for which
98\% of the runs find the problem optimum. A fixed-size selectorecombinative GA is used to search the minimum population size required using random initialization, to provide enough BBs to converge to the optimum using only recombination and selection mechanisms. 

Algorithm \ref{alg:bisection} depicts the method based on bisection.
The method begins with a small population size which is doubled until
the algorithm ensures a reliable convergence. We define the
reliability criterion as the convergence of the algorithm to the
optimum 49 out of 50 times (0.98 of Success Rate). After that, the
interval $(\mathit{min},\mathit{max})$ is halved several times and the population size
adjusted within such a range until $\frac{\mathit{max}-\mathit{min}}{\mathit{min}} > \mathit{threshold}$,
where $\mathit{min}$ and $\mathit{max}$ stand respectively for the minimum and maximum population size estimated and $\mathit{threshold}$ for the accuracy of the adjustment within such a range. This parameter has been set to $\frac{1}{16}$ in order to obtain a good adjustment of the initial population size.
\begin{algorithm}[tb]
\caption{Method based on bisection}
\label{alg:bisection}
\begin{algorithmic}
\STATE $n'$ = initial population size
\WHILE{reliability of $P_{init}'$ with size $n'$ $<$ 98\%}
\STATE $\mathit{min} = n'; \mathit{max} = 2n'; n' = 2n'$
\STATE $P_{init}'$ size $\leftarrow$ $n'$
\ENDWHILE
\WHILE{ $\frac{\mathit{max}-\mathit{min}}{\mathit{min}} > \frac{1}{16}$}
\STATE $n' = \frac{max+min}{2}$ 
\STATE $P_{init}'$ size $\leftarrow$ $n'$
\IF{reliability of $P_{init}'$ with size $n'$ $<$ 98\%}
\STATE $\mathit{min} = n'$
\ELSE
\STATE $\mathit{max} = n'$
\ENDIF
\ENDWHILE
\end{algorithmic}
\end{algorithm}

\subsection{Refining Initial Supply of Building Blocks}
\label{sec:refinement}

The initial supply of BBs given by a population $P_{init}'$ might be a bit oversized because the minimum population size $n'$ is estimated stochastically. Such a precision is usually accurate enough for scalability analysis, but in the case of study, additional BBs will induce smaller values for $\tau$ and $\rho$, leading to a slightly unfair comparison.

Therefore, Algorithm \ref{alg:refinement} shows an iterative process to refine $P_{init}'$ into $P_{init}$ by randomly subtracting 1\% of the individuals until $P_{init}$ is minimized.
\begin{algorithm}[tb]
\caption{Refinement of $P_{init}'$}
\label{alg:refinement}
\begin{algorithmic}
\STATE $P_{init}\leftarrow P_{init}'$
\WHILE{reliability of $P_{init}$ with size $n$ $\geq$ 98\%}
\STATE $P_{init} \leftarrow$ randomly removes 1\% of individuals in $P_{init}$
\ENDWHILE
\end{algorithmic}
\end{algorithm}

\subsection{Varying Population Size}
\label{sec:varying}

\begin{figure}
\centerline{\includegraphics[width=.6\linewidth]{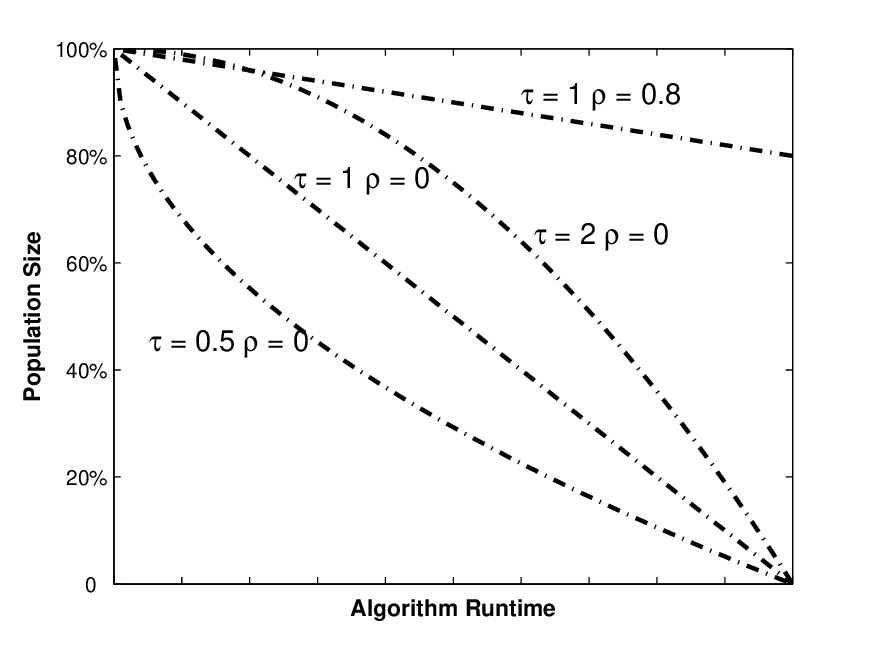}}
\caption{SVPS shapes for different values of $\tau$-$\rho$.}
\label{fig:equation}
\end{figure}

The SVPS-GA uses the following deterministic function to vary its size:
\begin{equation} \label{eq:distribution}
n_g = \left\{\begin{array}{lr}
  n_0 \times \left( 1- (1-\rho) \left(\frac{g}{g_{max}}\right)^{\tau} \right), & g\leq g_{max}\\
  n_{g_{max}}, & {g}>g_{max}
\end{array}\right.
\end{equation}
In this equation, $n_g$ stands for the population size at generation $g$, $n_0$ the initial population size, and $n_{g_{max}}$ the population size when a maximum number of generations ($g_{max}$) of the schedule is reached. To scale the shape of the function into the SVPS-GA runtime, $g_{max}$ is estimated on the necessary runtime of the fixed-size GA. $\tau$ and $\rho$ are respectively the speed and severity parameters. As shown in Figure \ref{fig:equation}, the values of $\tau$ and $\rho$ influence the shape of the population sizing schedule. 
A smaller $\tau$ leads to a faster reduction of the population size. $\rho$ belongs to the interval $]0,1]$, where a value of $\rho\rightarrow 0$ means that the run ends with a nearly empty population, and where a value of $\rho=1$ does not modify the initial population size.

Algorithm \ref{alg:varying} shows the procedure for iterating on different values of $\tau$ and $\rho$. 
\begin{algorithm}[tb]
\caption{Varying the population with $\tau$-$\rho$}
\label{alg:varying}
\begin{algorithmic}
\FOR{$\tau = 0.125,..._{*1.5},32$} 
\FOR{$\rho = 0.25,..._{+0.5},1$}
\IF{SVPS-GA reliability ($P_{init}$,$\tau$,$\rho$) $\geq$ 98\%}
\STATE store the combination $\tau$-$\rho$
\ENDIF
\ENDFOR
\ENDFOR
\end{algorithmic}
\end{algorithm}
Only those results which guarantee a success rate of 0.98 are stored (e.g. a very small $\tau$ and $\rho$ could lead to an unreliable convergence of the SVPS-GA). Lower bounds for $\tau$ and $\rho$ have been fixed to pessimistic values that will not meet the reliability condition (e.g. $\tau=0.125$ means that the population size will converge too quickly to $\rho$), while upper bounds stand for equivalent sizes to the fixed-size population (e.g. $\rho=1$ means that the population size does not shrink, or $\tau=32$ that the population shrinks in the last few generations).

\section{Experimental Setup}
\label{sec:setup}

\begin{figure}
\centerline{\includegraphics[width=.6\linewidth]{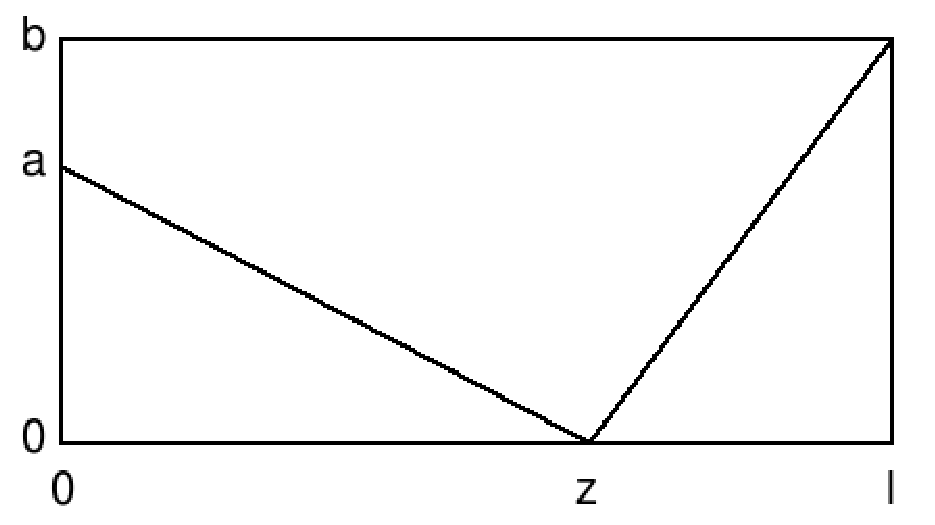}}
\caption{Generalized \emph{l-trap} function. }
\label{fig:trap}
\end{figure}

Following Lobo and Lima's recommendations \cite{lobo07:review} on the selection of a test suite with known population requirements and investigating the scalability on landscapes of different characteristics, experiments were conducted on trap functions \cite{ackley:trap}. A trap function is a piecewise-linear function defined on unitation (the number of one values in a binary string). There are two distinct regions in the search space, one leading to a global optimum and the other leading to the local optimum (see Figure \ref{fig:trap}).
In general, a trap function is defined by the following equation:
\begin{equation} \label{eq:trap}
trap(u(\overrightarrow{x}))=\left\{
\begin{array}{ll}
\frac{a}{z}(z-u(\overrightarrow{x})), & \mbox{if}~u(\overrightarrow{x})\in[0,z] \\
\frac{b}{l-z}(u(\overrightarrow{x})-z), & \mbox{if}~u(\overrightarrow{x})\in]z,l]
\end{array} \right.
\end{equation}
where $u(\overrightarrow{x})$ is the unitation function returning the number of one values in bit string $\overrightarrow{x}$, \textit{a}\ is the local optimum, \textit{b}\ is the global optimum, \textit{l}\ is the problem size and \textit{z}\ is a slope-change location separating the attraction basin of the two optima. 

For the following experiments, 2-trap, 3-trap and 4-trap functions were designed with the following parameter values: $a = l-1$, $b = l$, and $z = l-1$. With these settings, 2-trap  is not deceptive, 4-trap is deceptive and 3-trap lies in the region between deception and non-deception. Under these conditions, it is possible not only to examine the scalability on trap functions, but also to investigate how the scalability varies when changing from non-deceptive to deceptive search landscapes.
Scalability tests were performed by juxtaposing \textit{m}\ trap functions and summing the fitness of each sub-function to obtain the total fitness. 

All settings are summarized in Table \ref{table:parameters}, operators as binary tournament or one-point crossover are standard in GAs \cite{eiben:eas}. 
\begin{table}
\centering
{\small
\begin{tabular}{ll}
\multicolumn{2}{l}{\textbf{Trap instances}}\\
\hline
BB size ($l$) & $2, 3, 4$\\
Number of sub-functions ($m$) & $2, 4, 8, 16, 32, 64$\\
&\\
\multicolumn{2}{l}{\textbf{GA Setup}}\\
\hline
GA & selectorecombinative GA \\
& selectorecombinative SVPS-GA \\
Population size & Bisection + Refinement \\
Selection of Parents & Binary Tournament\\
Recombination & One-point crossover, $p_c = 1.0$ \\
Individual Length & $ l\times m $\\
&\\
\multicolumn{2}{l}{\textbf{SVPS Setup}}\\
\hline
Speed ($\tau$)    & $0.125,..._{*1.5},32$ \\
Severity ($\rho$) & $0.25,..._{+0.05},1$ \\
\end{tabular}
}
\caption{Parameters of the experiments}
\label{table:parameters}
\end{table}
The baseline for comparison is a generational selectorecombinative GA. Since the methodology imposes a 98\% success rate in the results, the Average Evaluations to Solution (AES) has been used as an appropriate metric to measure the computational effort to reach the success criterion. A more efficient algorithm requires a smaller number of evaluations.

\section{Results}
\label{sec:results}

\begin{figure*}
\centerline{\subfigure{\includegraphics[width=1.6in]{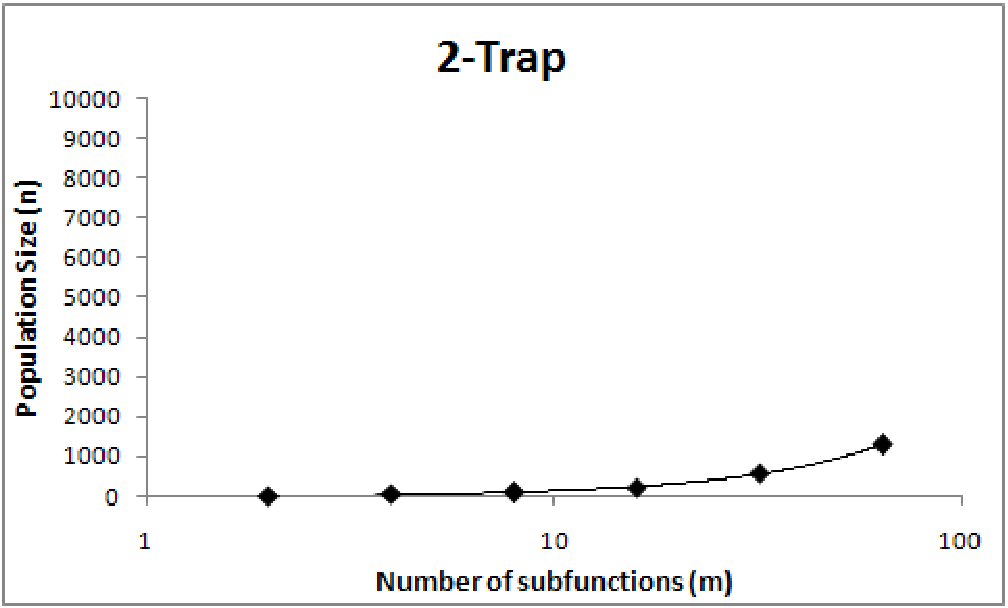}}
\hfil
\subfigure{\includegraphics[width=1.6in]{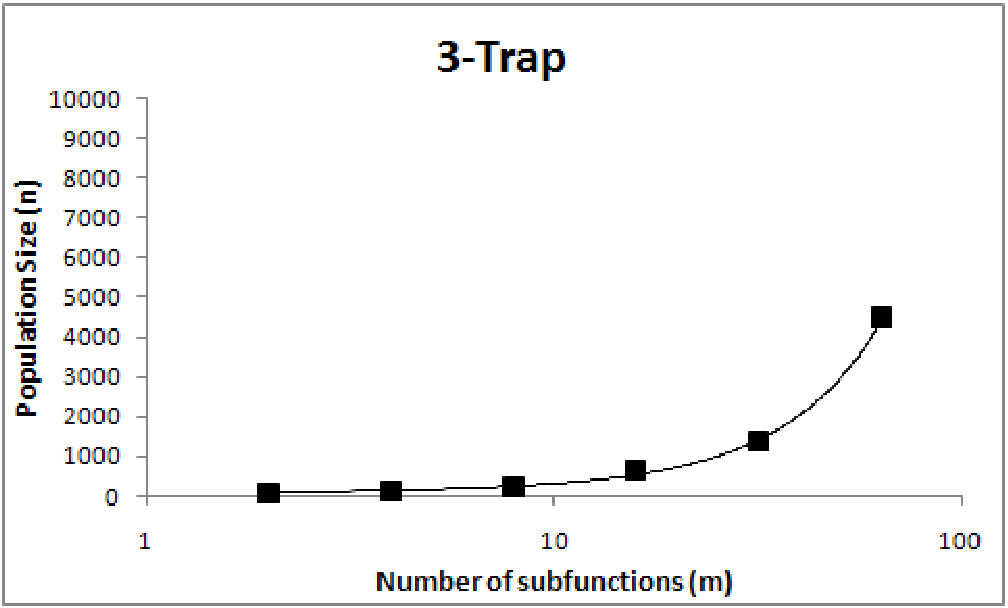}}
\hfil
\subfigure{\includegraphics[width=1.6in]{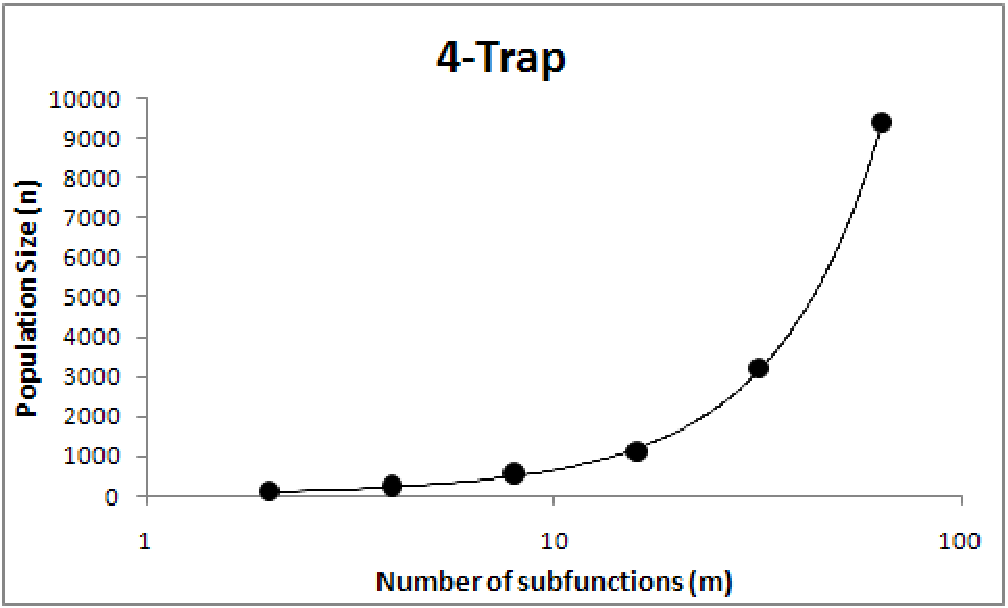}}
}
\centerline{\subfigure{\includegraphics[width=1.6in]{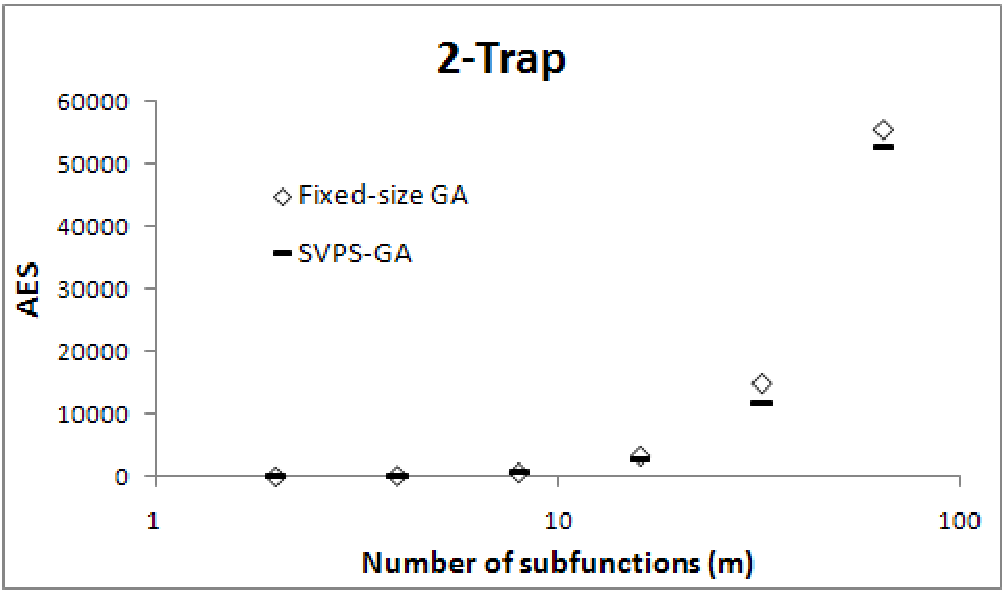}}
\hfil
\subfigure{\includegraphics[width=1.6in]{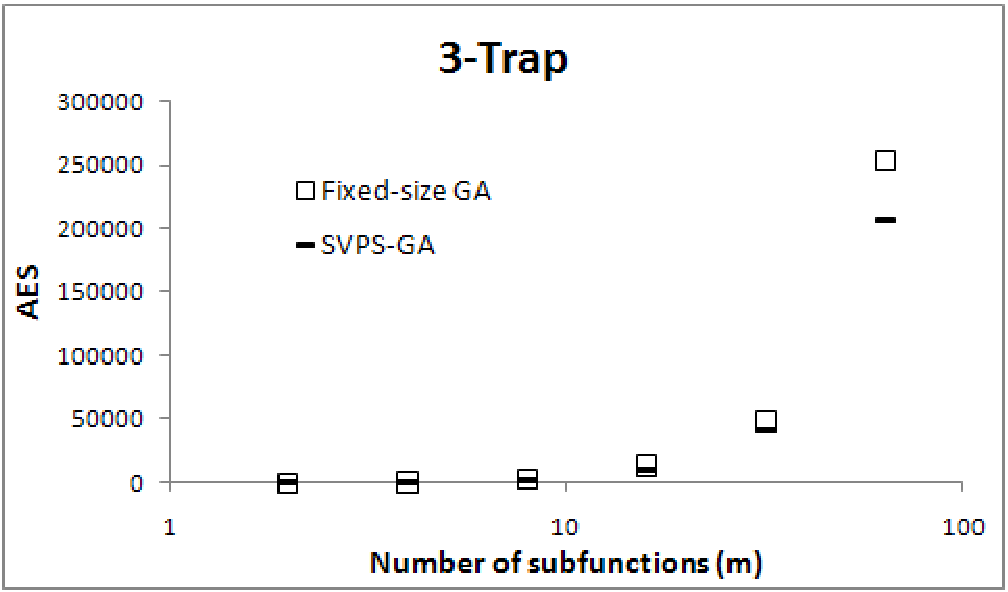}}
\hfil
\subfigure{\includegraphics[width=1.6in]{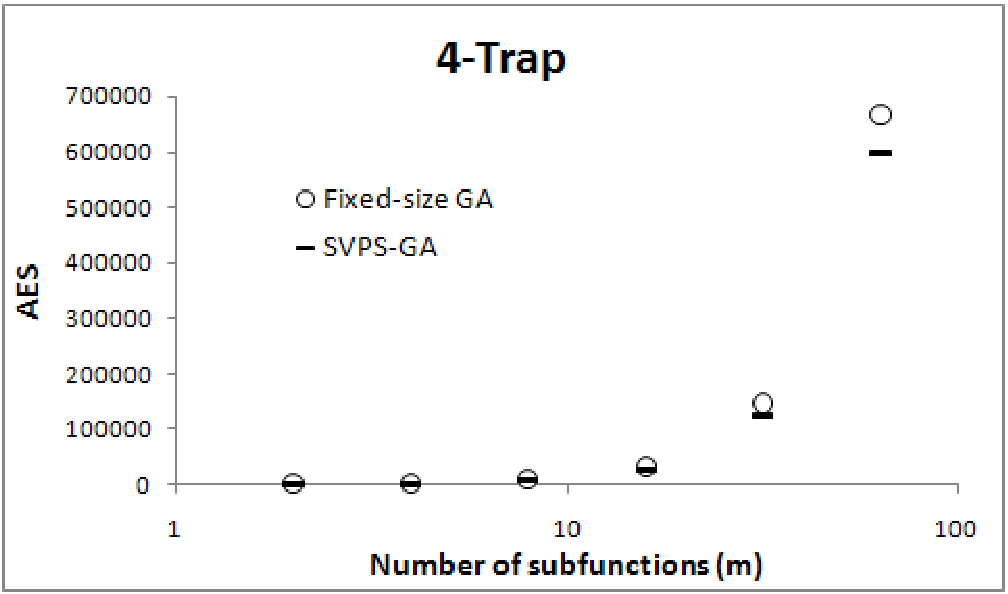}}
}
\caption{Scalability with trap functions. Optimal population size ({\tt top}) and Average Evaluations to Solution (AES) ({\tt bottom}) values for a fixed-size GA and the SVPS-GA. }
\label{fig:effort}
\end{figure*}

From the graphics in Figure \ref{fig:effort} it can be concluded that SVPS-GA scales better than the fixed-size GA. 
This fact proves that variable population sizing schemes can improve the performance of GAs.  Student's $t$-test conducted on the results in Table \ref{table:results} shows that, except for the smaller instances (i.e. some combinations for $\tau-\rho$ in $m= 2, 4$ and $8$), improvements are statistically significant.
\begin{figure}
\centerline{\includegraphics[width=.6\linewidth]{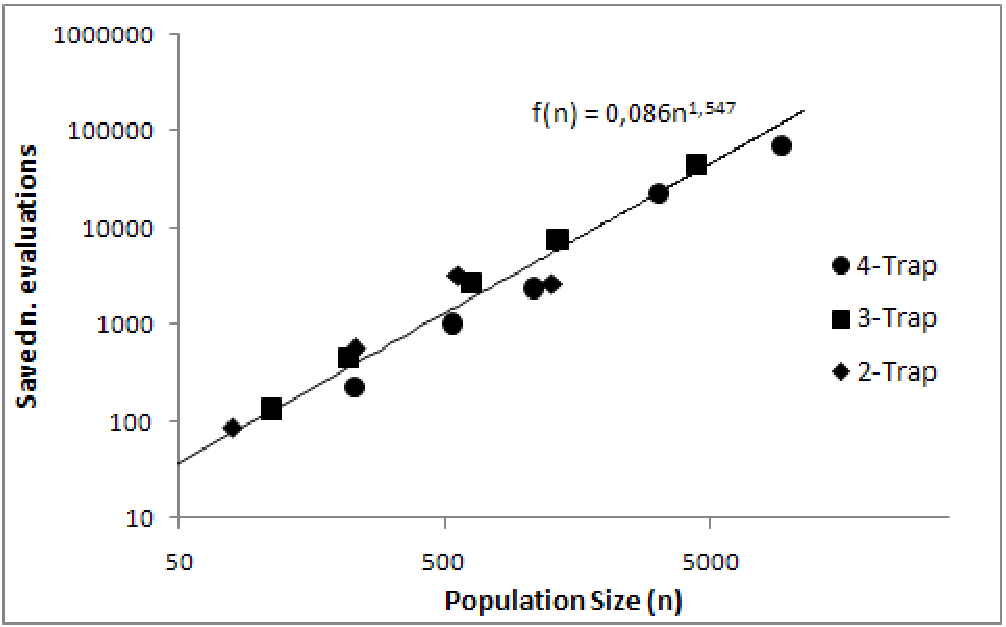}}
\caption{Improvement in the number of evaluations of the SVPS-GA with respect to the fixed-size GA. Results are depicted  as a function of the initial population size used by the different problem instances of 2-trap, 3-trap, and 4-trap functions.}  
\label{fig:savedeffort}
\end{figure}

\begin{table*}
\centering
  \rotatebox{90}{
{\tiny
\begin{tabular}{|c||c|c|c|c|c||c|c|c|c|c||c|c|c|c|c|}
\hline
&\multicolumn{5}{c||}{2-Trap}&\multicolumn{5}{c||}{3-Trap}&\multicolumn{5}{c|}{4-Trap}\\
\hline
&\multicolumn{2}{c|}{Fixed-size GA}&\multicolumn{3}{c||}{SVPS-GA}&\multicolumn{2}{c|}{Fixed-size GA}&\multicolumn{3}{c||}{SVPS-GA}&\multicolumn{2}{c|}{Fixed-size GA}&\multicolumn{3}{c|}{SVPS-GA}\\
\hline
\texttt{m}&\texttt{n}&AES&AES&$\tau$&$\rho$&\texttt{n}&AES&AES&$\tau$&$\rho$&\texttt{n}&AES&AES&$\tau$&$\rho$\\
\hline
2& 19.0 &30.02&\textbf{22.53}& 0.42& 0.25 &67&101.69&100.96&7.2&0.25&110&268.57&259.8&2.13& 0.25\\
&&$\pm$33.57&{\bf $\pm$8.17}& &  &&$\pm$49.76&$\pm$45&&&&$\pm$182&$\pm$152 & & \\
&&&{\bf 23.61}& 0.18& 0.5 &&&&&&&&264.02& 2.84& 0.3\\
&&&{\bf $\pm$11.47 }& &  && & &&&&   &$\pm$168  &&\\
&&&24.76 &0.28& 0.3 &&&&&&&&267.3 &0.125& 0.75\\
&& &$\pm$13.92 & &  &&& &&&&  &$\pm$161  &&\\

\hline
4& 49 &170&\textbf{144.02}&0.63 &0.7&112&597.9 &\textbf{467.04}&0.28 &0.7&230&1768.46 &\textbf{1549.14} &2.84& 0.7\\
&&$\pm$104 &{\bf $\pm$96.8} &&&&$\pm$221 &{\bf $\pm$136} &&&&$\pm$620 &{\bf $\pm$428} &&\\
&&&{\bf 145.56 }&0.28& 0.75&&&562.57 &3.2 &0.25&&&1636.63 &0.42& 0.95\\
&&&{\bf $\pm$64.5} &&&&& $\pm$188&&&&&$\pm$563 &&\\
&&&155.54 &0.9 & 0.65&&&581.06 &2.13&0.55&&&1677.57 &3.20 &0.65\\
&&&$\pm$83.4&&&&&$\pm$211 &&&&&$\pm$549 &&\\

\hline
8& 80 &668.49&\textbf{582.98} &7.2 &0.55&220 &2641.98&\textbf{2192.42} &0.9 &0.9&543&8714.10& \textbf{7713.2} &4.8 &0.6\\
&&$\pm$179.8 &{\bf $\pm$169} &&&&$\pm$648.5 &{\bf $\pm$381} &&&&$\pm$1952 &{\bf $\pm$1362} &&\\
&&&625.10 &10.81 &0.4&&&{\bf 2235.63 }&0.18 &0.95&&&{\bf  7822.95} &7.2 &0.35\\
&&&$\pm$194 &&&&&{\bf $\pm$363} &&&&&{\bf $\pm$1205} &&\\
&&&637.63 &24.32 &0.3&&&{\bf 2452.55} &7.2 &0.4&&& {\bf 7892.44} &0.9 &0.95\\
&&&$\pm$199 &&&&&{\bf $\pm$397} &&&&&{\bf $\pm$1687} &&\\
&&&&&&&&                {\bf 2471.52} &4.8 &0.65&&& {\bf 8223.72} &10.81 &0.25\\
&&&&&&&&{\bf $\pm$532} &&&&&{\bf $\pm$1153} &&\\

\hline
16& 230 &3355.57&\textbf{2804.08} &0.125 &0.95&639 &13451.8 & \textbf{10763.38} &0.42&0.95&1097&30721.04&\textbf{ 28384.51} &10.81 &0.55\\
&&$\pm$584.9 &{\bf $\pm$436 }&&&&$\pm$1861 &{\bf $\pm$1120 }&&&&$\pm$3348 &{\bf $\pm$ 2259}&&\\
&&&{\bf  2896.72 }&2.13 &0.8 &&&{\bf  11585.95} &4.8  &0.6 &&&{\bf  28602.94 }&2.13 &0.95\\
&&&{\bf $\pm$373 }&&&&&{\bf $\pm$ 988}&&&&&{\bf $\pm$2646 }&&\\
&&& {\bf 2984.48} &7.2&0.25&&&{\bf  11636.61} &7.2 &0.25&&&{\bf  29212.24} &16.21 &0.35\\
&&&{\bf $\pm$336 }&&&&&{\bf $\pm$891 }&&&&&{\bf $\pm$2348 }&&\\
&&& {\bf 3015.1} &4.8  &0.5 &&&{\bf  11684.92} &3.20  &0.8 &&&{\bf  29456.2} &7.2 &0.8\\
&&&{\bf $\pm$381 }&&&&&{\bf $\pm$923 }&&&&&{\bf $\pm$2674 }&&\\
&&&&&&&&                  {\bf  12023.58} &2.13  &0.9 &&& {\bf 29460.75} &24.32 &0.25\\
&&&&&&&&{\bf $\pm$1188 }&&&&&{\bf $\pm$2606 }&&\\

\hline
32& 568 &14943.96& \textbf{11661.64} &0.63 &0.95&1353 &48521.98& \textbf{40911.83} &4.8 &0.75&3225&146072.28 &\textbf{123758.85} &4.8 &0.8\\
&&$\pm$1475 &{\bf $\pm$875 }&&&&$\pm$5257 &{\bf $\pm$2980 }&&&&$\pm$10777 &{\bf $\pm$6228 }&&\\
&&& {\bf 12401.14} &3.20  &0.8 &&&{\bf  44037.66} &10.81 &0.6 &&&{\bf  127900.89} &7.2 &0.7\\
&&&{\bf $\pm$834 }&&&&&{\bf $\pm$2985 }&&&&&{\bf $\pm$6957 }&&\\
&&& {\bf 13223.79} &10.81 &0.35&&&{\bf  46072.4 } &16.21 &0.5 &&& {\bf 134020.22 }&16.21 &0.3\\
&&&{\bf $\pm$1032 }&&&&&{\bf $\pm$3342 }&&&&&{\bf $\pm$6396 }&&\\
&&& {\bf 13327.55} &7.2 &0.65&&& {\bf 46519.18} &24.32 &0.25&&&{\bf  134052.36 }&3.2 &0.95\\
&&&{\bf $\pm$1285 }&&&&&{\bf $\pm$2925 }&&&&&{\bf $\pm$9115 }&&\\
&&& {\bf 13802.71} &16.21 &0.25&&&&&&&& {\bf 138275.59} &24.32 &0.25\\
&&&{\bf $\pm$1102 }&&&&&&&&&&{\bf $\pm$7375 }&&\\

\hline
64& 1280 &55415.1 &\textbf{52773.36} &16.21 &0.7&4480 &252368.92& \textbf{207263.65} &3.2 &0.9&9408&666693.86&\textbf{596770.36} &10.81 &0.65\\
&&$\pm$3501 &{\bf $\pm$2408 }&&&&$\pm$13604 &{\bf $\pm$8435 }&&&&$\pm$37471 &{\bf $\pm$22940 }&&\\
&&&{\bf 52843.26 }&24.32 &0.25&&&{\bf  227992.93 }&10.81 &0.65&&&{\bf 619930.93 }&16.21 &0.55\\
&&&{\bf $\pm$2494 }&&&&&{\bf $\pm$9315 }&&&&&{\bf $\pm$25075 }&&\\
&&&{\bf 52963.61 }&10.81 &0.85&&&{\bf  229857.77 }&16.21 &0.3 &&&{\bf 632222.06} &24.32 &0.45\\
&&&{\bf $\pm$2843 }&&&&&{\bf $\pm$8830 }&&&&&{\bf $\pm$25380 }&&\\
&&&{\bf 54897.51 }&7.2 &0.95&&&&&&&&&&\\
&&&{\bf $\pm$3868 }&&&&&&&&&&&&\\
\hline
\end{tabular}
}}
\caption{Experimental results, values in bold are statistically significant.}
\label{table:results}
\end{table*}

Such improvements are more remarkable when the GA requires large population sizes which is directly related with the problem size and difficulty.
As a general pattern, the larger the number of sub-functions ($m$) or the more difficult the problem, the larger the initial population is. Hence, SVPS performs better under large instances of difficult problems. 

The relationship between the initial population size and the improvement in the number of evaluations is depicted in Figure \ref{fig:savedeffort}.
It shows the saved evaluations by the SVPS-GA  with respect to the fixed-size GA. The improvement keeps a proportionality of order $\Theta(n^{1.54})$ to the initial population size. However, it is our belief that other variable population sizing schemes could outperform this mark since we have just contemplated the shrinkage of the population.

\begin{figure*}
\centerline{\subfigure{\includegraphics[width=1.6in]{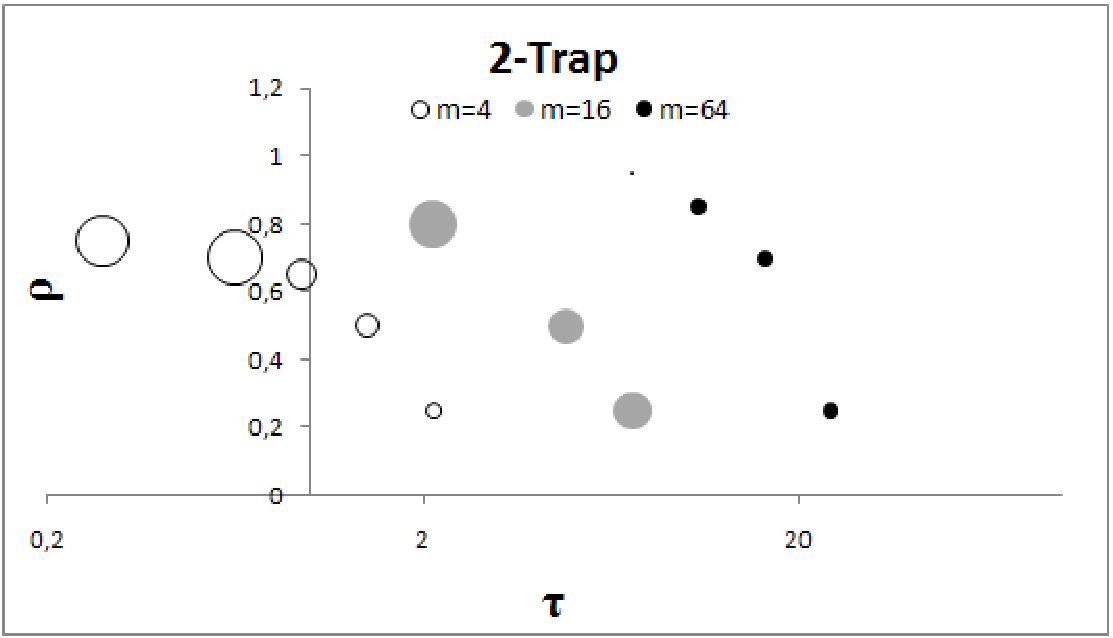}}
\hfil
\subfigure{\includegraphics[width=1.6in]{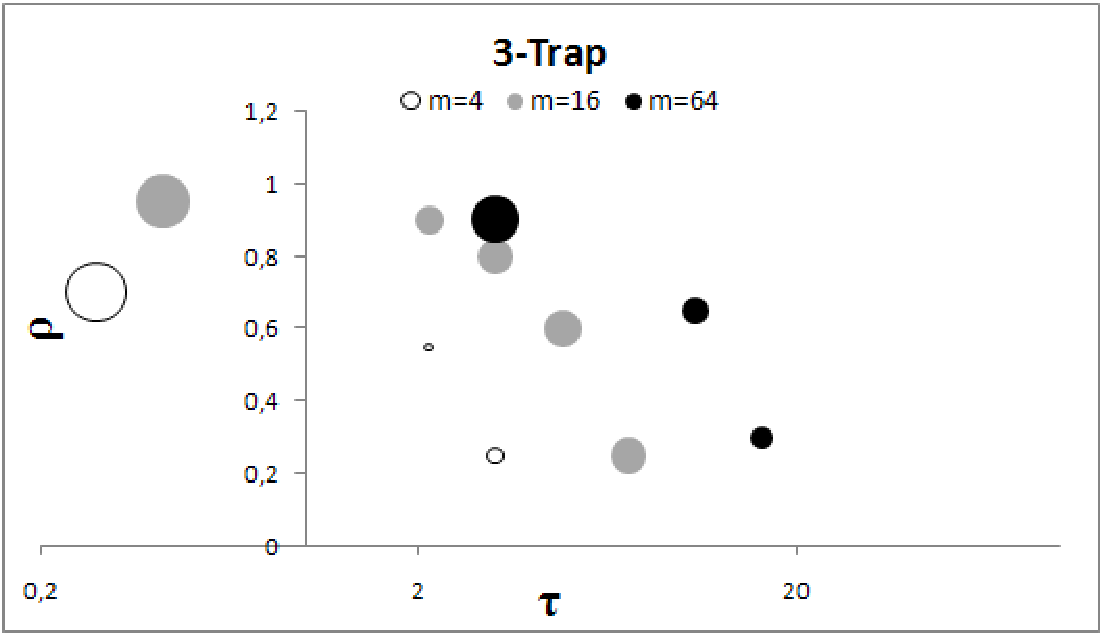}}
\hfil
\subfigure{\includegraphics[width=1.6in]{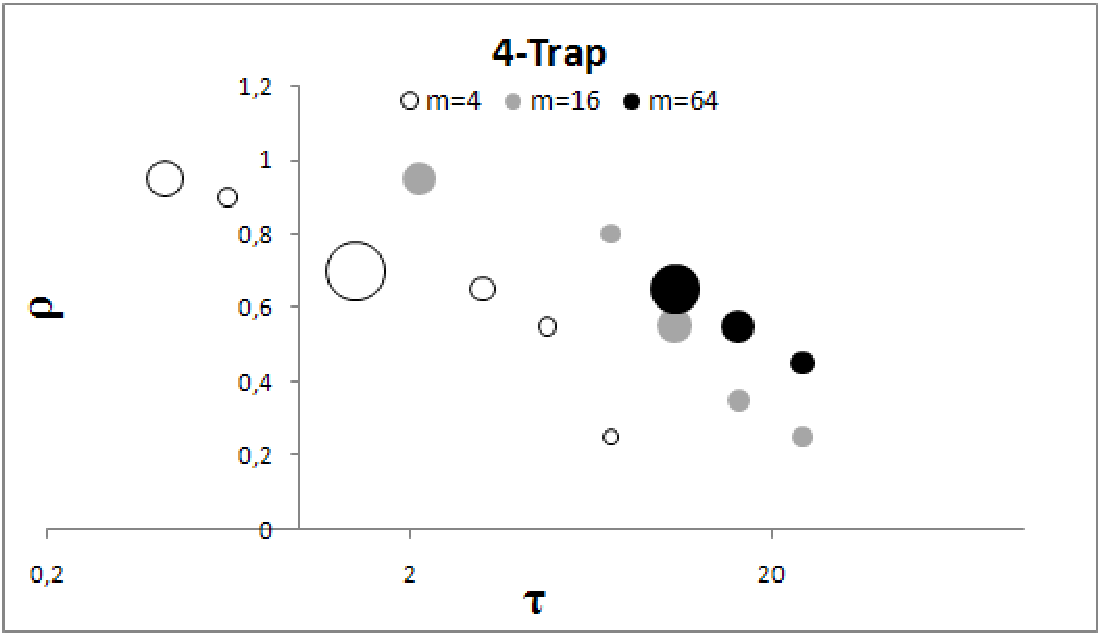}}
}
\caption{SVPS-GA combinations of $\tau$-$\rho$ converging to a SR of
 0.98. From left to right, 2-trap, 3-trap, and 4-trap are represented
 for sub-functions values of $m=4$, $m=16$, and $m=64$. The area of the
 circles stands for the AES improvement with respect to the fixed-size
 GA.}
\label{fig:taurho}
\end{figure*}

In order to gain some insight into the way the population shrinks, Figure \ref{fig:taurho} shows the set of strategies (i.e. combinations $\tau-\rho$) in which the SVPS-GA outperforms the fixed-size GA. 
They present the following behavior:
\begin{itemize}
\item For a given problem instance, values of $\rho$ decrease as values of $\tau$ increase.
\item Small values of $\tau$ usually report a better GA performance. 
\item As the problem scales, the set of strategies is shifted to higher values of $\tau$.
\end{itemize}

Therefore, as a good strategy, the GA tends to end with a high percentage of the initial population size ($\rho$) but losing individuals from a very early stage on the GA run ($\tau$). Nevertheless, large instances require that the population remains almost intact for a longer period, allowing a shrinkage mainly at the last stage of the GA run.

Despite the fact that SVPS has not been designed to find an optimal variability in the population size, it improves the GA performance. Such a result is significant since it proves that variable population sizing schemes are an open issue in the design of GAs.

\section{Conclusions}
\label{sec:conclusions}

In this paper we have presented a framework to test variable population sizing schemes. Additionally, a Simple Variable Population Sizing (SVPS) scheme is proposed in order to gain some insights into the population requirements for the different stages of a GA run. The framework consists in a three step methodology. The first and the second steps provide the initial population to be used for the variable population sizing scheme. This initial population represents the minimum initial supply of raw BBs that a fixed-size selectorecombinative GA needs to converge to the optimum solution. The third step is a deterministic resizing schedule configured by a speed and a severity parameter.

The framework has been designed to be compliant with the recommendations of Lobo and Lima \cite{lobo07:review} for the analysis of variable population sizing schemes:
\begin{itemize}
\item The test suite (i.e. trap functions) has known population sizing requirements.
\item A scalability analysis has been conducted by varying the size and the complexity of the problem instances.
\item The initial population size is well adjusted so that the GA will converge to the problem optima with a success rate of 0.98. The initial supply of BBs and the solution quality are fixed for both population sizing schemes, allowing a fair comparison based on the difference of the number of evaluations.
\end{itemize}

Preserving the condition of optimality (i.e. success rate of 0.98), the SVPS provides not a single but a set of strategies. Independently of the problem instance, the set follows a common pattern: a large value of $\tau$ implies a small one of $\rho$, that is, the population shrinkage strategy has to keep a balance between severity and speed of the population shrinkage. Results show that SVPS requires a smaller number of evaluations than the fixed population sizing scheme, and therefore, improves the GA performance. 

Future work will include studying how the addition of variation
operators such as mutation affect performance and its scaling, and
also fine-tuning which values of $\rho$ and $\tau$ are the most
appropriate for a wide range of applications.

\section *{Acknowledgements}
This work has been supported by the Spanish MICYT project
TIN2007-68083-C02-01, the Junta de Andalucia CICE project P06-TIC-02025,
and the Granada University PIUGR 9/11/06 project. The authors 
are grateful to Annette Schwerdtfeger for proofreading this manuscript.

\vfill

\bibliographystyle{abbrv}

\end{document}